\documentclass{article}

\PassOptionsToPackage{numbers, compress}{natbib}
\usepackage[preprint]{neurips_2026}

\usepackage{microtype}
\usepackage{graphicx}
\graphicspath{{pictures/}}
\usepackage{subcaption}
\usepackage{booktabs}
\usepackage{caption}
\usepackage{hyperref}
\usepackage{url}
\usepackage{amsmath}
\usepackage{amssymb}
\usepackage{amsfonts}
\usepackage{mathtools}
\usepackage{amsthm}
\usepackage{enumitem}
\usepackage{xspace}
\usepackage{multirow}
\usepackage[table]{xcolor}
\usepackage{nicefrac}
\usepackage{algorithm}
\usepackage{algorithmic}
\usepackage[capitalize,noabbrev]{cleveref}

\theoremstyle{plain}

\theoremstyle{definition}

\theoremstyle{remark}

\def\modelname{ATLAS\xspace}

\title{Think, Plan, Paint: Layout-Aware Reasoning for Controllable Image Generation in Unified Models}

\author{%
  Junhao Liu\thanks{\texttt{email: liujunhao@pku.edu.cn}} \\
  Hunyuan, Tencent \\
  School of Computer Science, Peking University \\
  \And
  Jian-Wei Zhang \\
  Hunyuan, Tencent \\
  \And
  Tao Huang \\
  Hunyuan, Tencent \\
  \And
  Miles Yang \\
  Hunyuan, Tencent \\
  \And
  Zhao Zhong\thanks{Corresponding author.} \\
  Hunyuan, Tencent \\
  \And
  Liefeng Bo \\
  Hunyuan, Tencent \\
}

\begin{document}

\maketitle

\begin{abstract}
Unified Multimodal Large Language Models (MLLMs) offer a promising paradigm for unifying visual understanding and generation, yet they still struggle to follow complex spatial instructions and logical constraints in controllable image generation. 
To address this gap, we present ATLAS, a unified framework that equips MLLMs with a human-like ``Think, Plan, and Paint'' paradigm.  
We adopt layout as the shared representation that connects the three stages, enabling the model to reason about spatial requirements, plan explicit object arrangements, and render the final image.
We further improve plan-to-image fidelity with reinforcement-learning-based layout alignment.
We instantiate ATLAS at 7B and 80B scales, achieving state-of-the-art performance among MLLMs on image generation benchmarks and an average 65.31\% improvement over existing layout-based unified MLLMs. On spatially related tasks, ATLAS obtains an average 23.06\% gain over the base models. Through the same layout interface, ATLAS also supports instruction-guided editing and multimodal grounding. We further introduce ATLAS-Reasoning, a benchmark for evaluating generation under complex spatial instructions.
\end{abstract}
\begin{figure}[t]
  \centering
  \includegraphics[width=\textwidth]{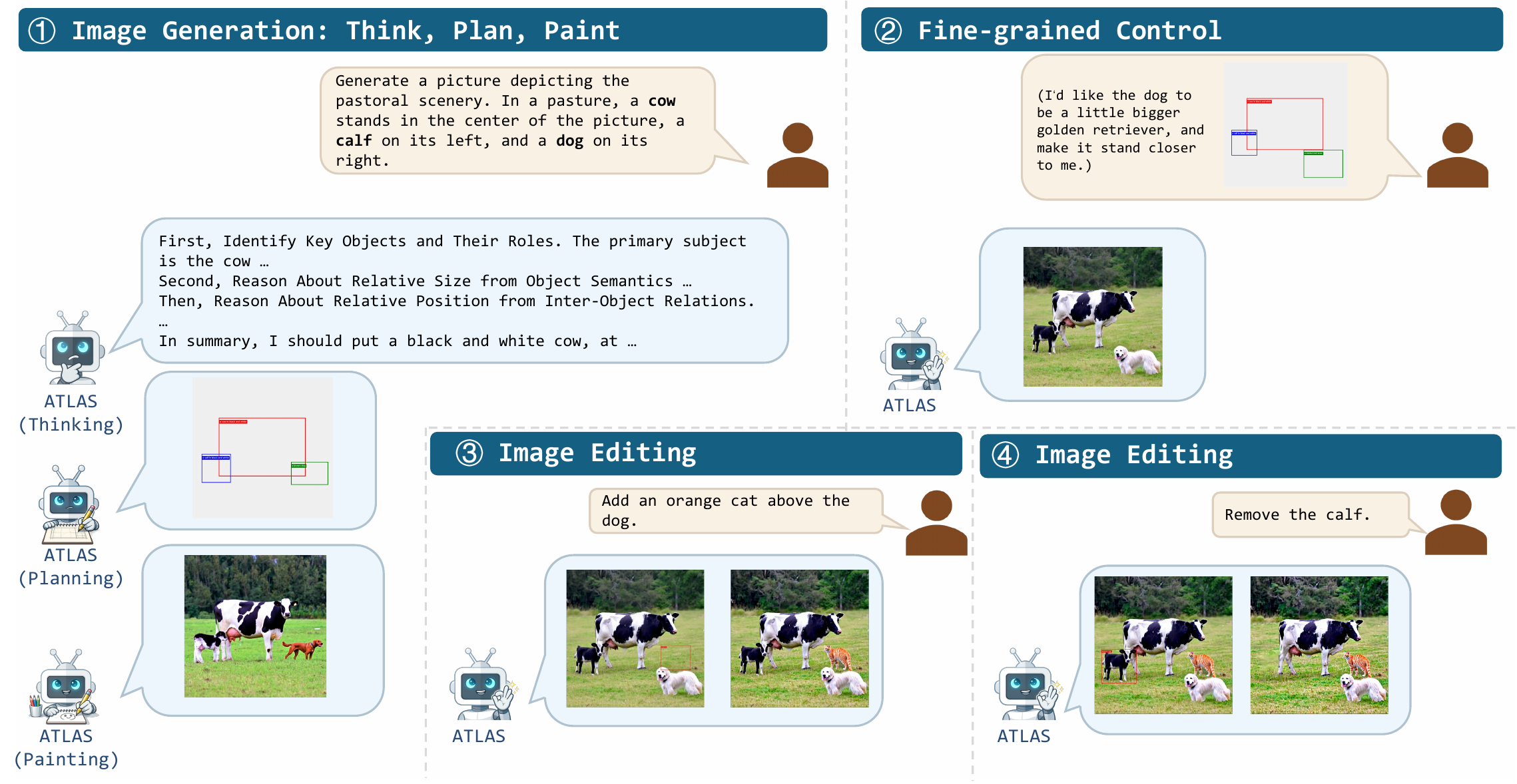}
  \caption{Our \modelname~framework generates images following a human-like \texorpdfstring{\textit{Think, Plan, and Paint}}{Think, Plan, and Paint} paradigm, and supports interleaved image generating and editing. Additionally, the Think and Plan phases provide the user with fine-grained control over the generation process.}
  \label{fig:teaser}
\end{figure}

\section{Introduction}

In image generation, a promising direction is to introduce \textit{Layout} as an intermediate representation to bridge the gap between abstract textual instructions and pixel-level arrangements, enabling more controllable image generation~\citep{gldesigner, layercraft, layoutgpt}.
Meanwhile, unified Multimodal Large Language Models (MLLMs) have emerged as a promising paradigm for image understanding and generation~\citep{janus, bagel,hunyuanimage30}.  
\citet{plangen} have taken an initial step toward integrating layout planning and image generation into unified MLLMs.
However, this line of work faces two critical limitations in practice.
First, layout planning is often performed as a direct prediction from the prompt, without an explicit reasoning process to decompose objects, counts, and spatial relations, which can lead to incomplete or inconsistent layouts under complex instructions.
Second, layout is still treated mainly as an auxiliary control signal for generation, rather than a shared spatial representation that can consistently support  planning, generation,  editing, and understanding.

To address these challenges, we argue that the potential of unified MLLMs lies in treating layout as a cross-modal bridge rather than a mere control signal, which enables the model to leverage the reasoning strengths of the text modality to structure the visual modality.
In this view, the explicit layout representation serves as a transparent window into the generation process, converting the "black-box" workflow into an interpretable one, and functions as an interactive interface that empowers users to exert fine-grained control over the final synthesis.

Motivated by this view, we propose \textbf{ATLAS} (\textbf{A}ligned \textbf{T}hinking \textbf{L}ayout-\textbf{A}ware image \textbf{S}ynthesis), a framework that enables unified MLLMs to generate images through a human-like \textbf{\textit{``Think, Plan, Paint''}} paradigm as shown in Figure~\ref{fig:teaser}.

Specifically, ATLAS initiates a \textbf{\textit{Think}} process to analyze object semantics and spatial constraints before executing the \textbf{\textit{Plan}} phase. To operationalize the cross-modal bridge role of layout, we introduce Shared Positional Tokens as a unified spatial vocabulary across understanding, planning, generation, and editing, and further apply Reinforcement Learning (RL)-based layout alignment to the \textbf{\textit{Paint}} stage so generated objects better align with the planned layout.
These explicit intermediate states make generation more interpretable and editable, while the Transfusion-style \textbf{\textit{Paint}} stage turns the planned layout into high-fidelity images.

We implement ATLAS on two unified auto-regressive Transfusion-style backbones~\citep{bagel,hunyuanimage30} and verify the effectiveness and scalability of our approach by instantiating ATLAS at two scales: \textbf{7B} and \textbf{80B}. Our experiments demonstrate that ATLAS outperforms existing layout-based unified MLLMs. In particular, our \textbf{80B} model achieves comparable generation performance to leading closed-source large models and outperforms existing layout-based unified MLLMs.

In summary, our contributions are:
\begin{itemize}[leftmargin=*,itemsep=0pt,topsep=0pt,parsep=0pt]
    \item We propose \textbf{ATLAS}, a unified MLLM framework with the \textbf{\textit{``Think, Plan, Paint''}} paradigm. We leverage the reasoning strengths of the text modality to structure the visual modality, enhancing the capability to handle complex spatial instructions while enabling precise, fine-grained user editing.
    \item We empower the \textbf{\textit{Think}} and \textbf{\textit{Plan}} phases with precise spatial grounding to ensure logical validity. By establishing a unified spatial vocabulary via Shared Positional Tokens and further aligning the \textbf{\textit{Paint}} stage with RL-based layout alignment, we improve the consistency between coherent plans and generated objects.
    \item We instantiate \modelname at \textbf{7B} and \textbf{80B} scales. ATLAS outperforms existing layout-based unified MLLMs, and the \textbf{80B} model achieves comparable generation performance to leading closed-source large models and outperforms existing layout-based unified MLLMs.
    \item We introduce \textbf{ATLAS-Reasoning}, a new benchmark designed to rigorously evaluate reasoning-aware generation and layout planning under complex spatial and logical instructions.
\end{itemize}
\begin{figure*}
    \centering
    \includegraphics[width=\textwidth]{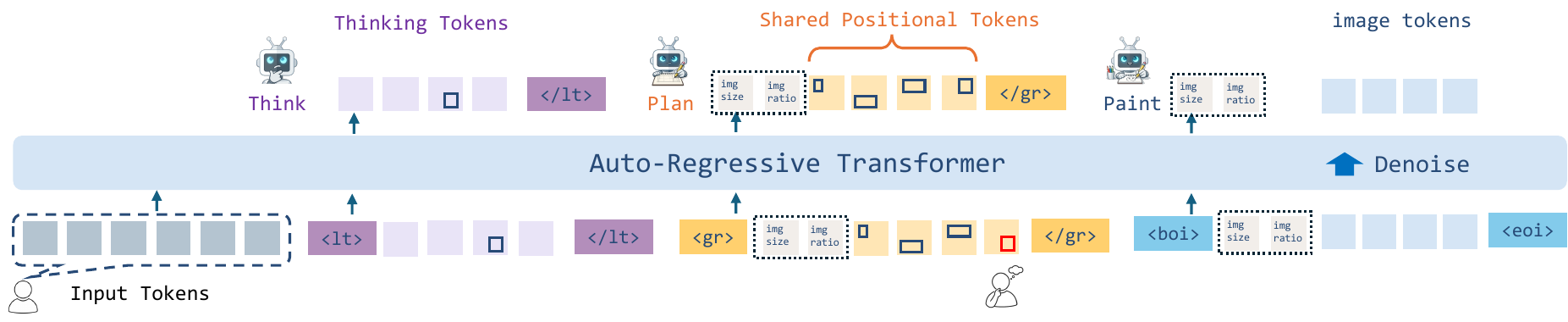}
    \caption{Overview of the \modelname~framework. Given a text prompt, \modelname~\textit{Thinks} through object semantics and spatial relations, \textit{Plans} a layout with Shared Positional Tokens that align text and vision and allow precise user edits, then \textit{Paints} a final image that faithfully follows the planned composition. Shared Positional Tokens represent object layouts in both \textit{Think} and \textit{Plan} phases.}
    \label{fig:method_overview}
\end{figure*}

\section{Background and Related Work}

\subsection{Unified Multimodal Large Language Models}
Unified Multimodal Large Language Models (MLLMs) are a mainstream paradigm for multimodal tasks.
While early models\citep{llava, qwenvl, deepseekvl, flamingo} focused primarily on understanding, recent works\citep{emu,chameleon,janus, seedx,luminamgpt} have unified input and output modalities, enabling arbitrary-to-arbitrary generation. A critical architectural evolution in this domain is the transition from discrete vector quantization to continuous representations. While discrete models often struggle with visual fidelity, recent hybrid architectures integrate AR-transformers with diffusion models~\citep{transfusion, bagel,hunyuanimage30,showo}. 

\subsection{Controllable Image Generation and Manipulation}
The pursuit of precise control in visual synthesis has evolved beyond simple text-to-image mapping. Early approaches focused on image manipulation, using text instructions to edit existing images~\citep{instructpix2pix, p2p, mgie}, or employing dense conditions (e.g., depth maps, canny edges) for structural control~\citep{controlnet, t2iadapter}. Among these interfaces, \textbf{Layouts} (e.g., bounding boxes) stand out as the most intuitive and explicit representation for defining spatial composition. Current layout-driven approaches generally fall into three paradigms:

\textbf{Adapter-based methods}~\citep{controlnet, gligen,Chen_2024_WACV, Xie_2023_ICCV} inject spatial conditions into frozen diffusion models. While effective, these methods are inherently \textit{passive}, requiring pre-defined layouts and lacking the intrinsic capability to ``plan'' from text alone. 

\textbf{Pipeline approaches}~\citep{vpgen,layoutgpt,layercraft, wang2024div} address this by using external LLMs to infer layouts for downstream generators. However, this creates a \textit{disjointed} system where planning is isolated from visual perception, leading to error propagation. 

\textbf{Unified approaches} like PlanGen~\citep{plangen} step forward by integrating planning and generation within a single model. 

Yet, as discussed below, they often treat layout generation as a direct translation task, overlooking the necessity of explicit reasoning, thus limiting their ability mainly to layout-guided generation, while struggling with layout planning.

\section{The \modelname Framework}
\label{sec:method}

\begin{figure*}[t]
    \centering
    \includegraphics[width=0.9\textwidth]{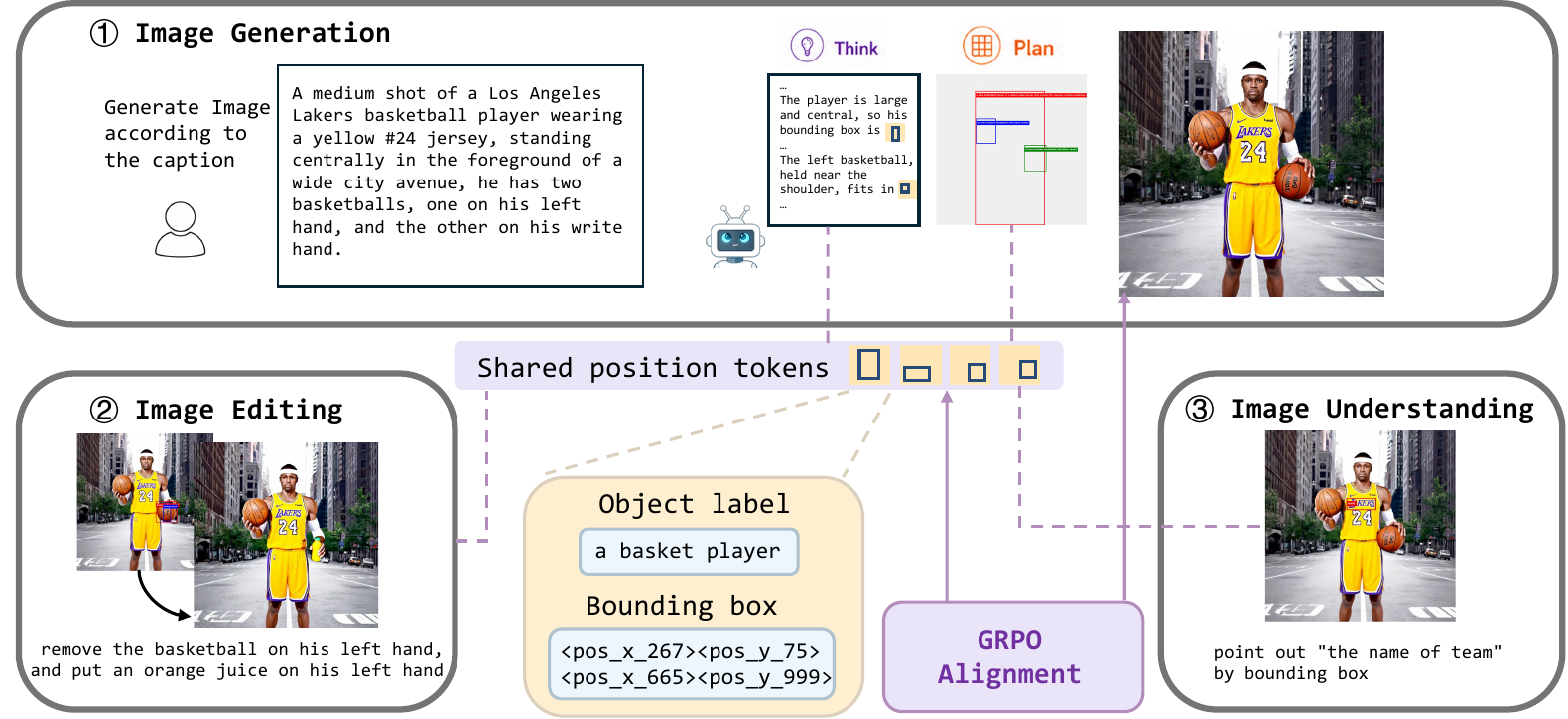}
    \caption{\modelname represents each grounded object with a semantic label followed by four dedicated coordinate tokens from the shared vocabulary $V_{pos}$.
The same label-box format serves as a unified spatial interface across image generation, image editing, and multimodal understanding.
We further apply reinforcement learning to align the planned layout with the generated image.}
\label{fig:shared_token}
\end{figure*}

In this section, we introduce \modelname in detail.
Figure~\ref{fig:method_overview} and Figure~\ref{fig:shared_token} illustrate the overall architecture and workflow of \modelname.

\subsection{Unified Architecture}
\subsubsection{Auto-regressive Transformer Backbone}
We instantiate ATLAS on two auto-regressive MLLM backbones, \textbf{BAGEL}~\citep{bagel} and \textbf{Hunyuan Image 3.0}~\citep{hunyuanimage30}.
Both models represent text and image signals in a shared sequence, where text is encoded as discrete embeddings and images are processed as continuous latent variables.
Using these two backbones allows us to evaluate ATLAS across different architectures and parameter scales.

\subsubsection{Shared Positional Tokens}
\label{sec:shared_tokens}
As illustrated in Figure~\ref{fig:shared_token}, we introduce \textbf{Shared Positional Tokens} as a unified spatial interface for image generation, image editing, and multimodal understanding.
Each grounded object is represented as a label-box pair $(c_i, b_i)$, where $c_i$ is the object label and $b_i=(x_i^{1}, y_i^{1}, x_i^{2}, y_i^{2})$ denotes the top-left and bottom-right corners of its bounding box.
We use a dedicated coordinate-token vocabulary $V_{pos}$ for spatial positions.
Specifically, each normalized coordinate $u\in[0,1]$ is assigned to one of 1000 position indices by $q(u)=\min(\lfloor 1000u \rfloor, 999)$, and we define $V_{pos}=V_x\cup V_y$, where $V_x=\{\texttt{<pos\_x\_0>},\ldots,\texttt{<pos\_x\_999>}\}$ and $V_y=\{\texttt{<pos\_y\_0>},\ldots,\texttt{<pos\_y\_999>}\}$.
Let $\tau_x(k)$ and $\tau_y(k)$ denote the corresponding $x$- and $y$-axis tokens in $V_{pos}$.
A box is then expressed as $\tau_x(q(x_i^{1})), \tau_y(q(y_i^{1})), \tau_x(q(x_i^{2})), \tau_y(q(y_i^{2}))$, such as \texttt{<pos\_x\_267><pos\_y\_75><pos\_x\_665><pos\_y\_999>} in Figure~\ref{fig:shared_token}.

\subsection{The ``Think, Plan, Paint'' Paradigm}
To fully leverage unified MLLM's capabilities, we structure the generation process into three explicit phases, transforming layout generation from an intuitive mapping into a reasoned deduction.

\subsubsection{Think: Layout-Aware Chain-of-Thought}
Before committing to any spatial arrangement, \modelname initiates a \textbf{Think} phase. 
Instead of blindly mapping text to coordinates, the model decomposes complex scenes into individual entities, deduces implicit spatial constraints with the capabilities of textual reasoning, and also takes visual priors into account by leveraging the Shared Positional Tokens.
This Layout-Aware Chain-of-Thought (CoT) serves as a cognitive foundation, enhancing the model's ability to plan coherent layouts. 

\subsubsection{Plan: Grounded Layout Generation}
Conditioned on the reasoning chain $R$, the model enters the \textbf{Plan} phase and generates a layout sequence $L=\{(c_i,b_i)\}_{i=1}^{n}$, where each pair follows the label-box token format defined in Section~\ref{sec:shared_tokens}.
Rather than directly predicting coordinates from the prompt alone, this phase uses the preceding reasoning to determine which objects should be grounded and how their boxes should satisfy the inferred spatial relations, counts, and attributes.
The resulting layout $L$ serves as an executable scene plan that can be inspected or edited by users and directly consumed by the subsequent \textit{Paint} phase.

\subsubsection{Paint: Conditioned Image Synthesis}
In the \textbf{Paint} phase, the model synthesizes the image $I$. The backbone denoises the random latent noise conditioned on the entire preceding sequence: the prompt $T$, the reasoning $R$, and the planned layout $L$. By attending to the precise layout tokens generated in the Plan phase, the model ``paints'' pixels that faithfully respect the specified spatial structure.

\subsection{Data Construction and Training Objectives}
To equip the model with this unified paradigm, we construct a specialized instruction-tuning dataset and employ a joint optimization strategy.

\subsubsection{Data Construction}
\label{sec:data_construction}

To empower \modelname with the ``Think, Plan, Paint'' capability, we curate a high-quality dataset centered around quadruples of $(T, R, L, I_{tgt})$, representing the Prompt, Reasoning, Layout, and target Image, respectively. The construction process focuses on establishing a logical causal chain:

\textbf{Logic-Driven Layout Annotation ($L$).} Unlike existing datasets~\citep{creatilayout, layoutgpt} that rely on indiscriminate object detection, we aim to train the model to plan layouts that are \textit{semantically deducible} from the text. We employ Qwen3-VL-235B-A22B~\citep{qwen3vl} to analyze both the text prompt $T$ and the target image $I_{tgt}$. We prompt the model to identify and bound only those objects required or implied by the prompt, filtering out irrelevant background. This ensures the layout $L$ serves as a precise execution plan for the text.

\textbf{Reasoning Chain Synthesis ($R$).} To bridge the gap between the abstract prompt $T$ and the concrete layout $L$, we synthesize the "Thinking" process. We prompt Qwen3-VL-235B-A22B to act as a spatial reasoner: given $T$ and the annotated $L$, it generates a reasoning chain $R$ that explicitly articulates the decomposition logic, spatial constraints, and attribute determinations that lead to the final layout. Additionally, we use an LLM to reject reasoning that is inconsistent with the layout or contains logical errors, ensuring high-quality supervision for the Think phase.

\textbf{Human-Verified Quality Control.}
To mitigate bias from synthesized reasoning data and LLM-based filtering, we iteratively refine the VLM/LLM prompts through manual inspection and verify the final data quality with an author audit.
Two authors check 500 layout annotations and confirm 95.3\% grounding accuracy for main objects.
They further check 200 synthesized reasoning chains and confirm 93.5\% reasoning accuracy, judging whether each chain is logically consistent with the prompt and the annotated layout.

\textbf{Data Diversity and Task Mixture.} To ensure robustness and prevent the model from over-relying on the full reasoning pipeline, we incorporate diverse data formats:
\begin{itemize}[leftmargin=15pt,itemsep=0pt,topsep=0pt,parsep=0pt]
    \item \textbf{``Think, Plan, Paint"  Data $(T, R, L, I_{tgt})$:} For learning complex spatial reasoning.
    \item \textbf{Layout-Only Data $(T, L, I_{tgt})$:} Standard grounded generation data without explicit reasoning.
    \item \textbf{Image-Only Data $(T, I_{tgt})$:} Large-scale text-to-image pairs to maintain general visual synthesis quality.
\end{itemize}
Finally, we conduct \textbf{joint training} on a mixture of these generation samples along with layout-aware Multimodal Understanding data $(I_{src},T,L)$ and Image Editing Data $(I_{src},T,L,I_{tgt})$. 

\subsubsection{Joint Optimization}
We train \modelname using a multi-task objective that combines language modeling and diffusion training.
For the discrete component (reasoning and layout tokens), we minimize the standard negative log-likelihood loss $\mathcal{L}_{txt}$. For the continuous component (image latents), we employ the diffusion loss $\mathcal{L}_{diff}$ (mse on velocity prediction). The total objective is formulated as:
\begin{equation}
    \mathcal{L} = \mathcal{L}_{txt}(R, L | T, I_{src}) + \lambda \mathcal{L}_{diff}(I_{tgt} | T, R, L, I_{src})
\end{equation}

The weight $\lambda$ balances the two losses. We set $\lambda$ following the training strategies of our base models.

\subsubsection{RL-based Layout Alignment}
After supervised joint training, we further refine the \textbf{Paint} stage with Flow-GRPO~\citep{flowgrpo}.
For each condition $(T,L)$ or $(T,R,L)$, the model samples a group of images and optimizes a group-relative objective.
We use a weighted sum of five rewards: layout position, prompt-level semantic alignment, region counting, negative-region suppression, and image quality.
For example, the layout-position reward evaluates whether generated objects appear in the specified boxes by matching detected objects with the input layout using semantic labels and spatial overlap.
After matching, we compute it as an IoU-based F1 score, $r_{\mathrm{pos}} = \frac{2 \cdot \mathrm{Precision} \cdot \mathrm{Recall}}{\mathrm{Precision} + \mathrm{Recall} + \epsilon}$.
Detailed definitions are provided in Appendix~\ref{appendix:rl_rewards}, and the contribution of this stage is evaluated in the ablation experiments.

\begin{table*}[t]
\centering
\caption{Quantitative comparison on T2I-CompBench and \modelname-Reasoning. We report the scores across six dimensions for T2I-CompBench and the \modelname-Reasoning benchmark. \modelname demonstrates comprehensive superiority, particularly in complex spatial reasoning and non-spatial logic attributes. For each metric, higher is better. The best results are \textbf{bolded}.}
\label{tab:main_results}
\resizebox{\textwidth}{!}{%
\begin{tabular}{l cccccc c}
\toprule
\multirow{2}{*}{\textbf{Method}} & \multicolumn{6}{c}{\textbf{T2I-CompBench}} & \multicolumn{1}{c}{\textbf{\modelname-Reasoning}} \\ 
\cmidrule(lr){2-7} \cmidrule(lr){8-8}
 & Color & Shape & Texture &  Spatial & Non-Spa. & Numeracy & Spatial \\ \midrule

\multicolumn{8}{l}{\textit{General T2I Models}} \\
SDXL~\citep{sdxl} &69.44 & 64.11&79.90&68.33&89.16 &	48.33 & 47.21	 \\
SD v3.5~\citep{sd3} & 90.37 &	80.37 & 93.12 & 80.83 & 91.66 & 17.43 & 54.17 \\
FluX.2-Dev~\citep{flux-2-2025} & 97.22 & 90.74 & 97.61 & 97.56 & 82.53 & \textbf{91.25} & 89.72 \\
Nano Banana Pro~\citep{gemini25} & 94.44 & 88.70 & \textbf{97.81} & 98.53 & \textbf{98.33} &87.12 & \textbf{95.31} \\
GPT-Image-1~\citep{gpt4o} & 95.74 &\textbf{92.31}& 97.78 &94.17 &96.67 &85.35  &93.48 \\
Show-O2~\citep{showo2} & 84.44 & 77.41	& 81.96 & 77.50	& 92.53	& 69.16 & 51.27\\
BAGEL  & 91.11 &81.64 & 93.12 & 78.33 & 90.50 & 70.83 & 52.01\\
BAGEL (Think) & 89.42 & 79.63 & 92.81 & 81.95 & 91.31 & 67.14 & 58.94 \\
Hunyuan Image 3.0 & 84.81 & 85.28 & 91.30 & 83.33 & 97.50 & 66.67 & 69.41 \\
Hunyuan Image 3.0 (Think) & 87.93 & 88.70 & 92.77 & 88.12 & 91.67 & 73.43 & 68.77 \\
\midrule

\multicolumn{8}{l}{\textit{Layout Pipeline/Agents}} \\
LayoutGPT~\citep{layoutgpt} &34.16 & 47.13 & 43.10 & 34.41 &46.59 &47.93 & 21.98 \\
GenArtist~\citep{genartist} & 89.43 & 87.17 & 84.32 & 95.31 & 92.56 & 84.41	& 88.29 \\ \midrule

\multicolumn{8}{l}{\textit{Layout-Based Unified MLLMs}} \\
PlanGen~\citep{plangen} &67.41 & 61.28 &77.00 &46.67 & 74.17 & 40.83 &25.43 \\ 

\rowcolor{gray!10}\modelname-7B & \textbf{97.13} & 86.17 & 96.26 & 86.01 & 93.79 & 83.41 & 72.31 \\
\rowcolor{gray!10}\modelname-80B & 96.42 & 90.16& 90.62 & \textbf{99.17} & 96.67 & 85.08 & 91.21 \\ \bottomrule
\end{tabular}%
}
\end{table*}

\section{Empirical Experiments}

\textbf{Implementation Details} We instantiate \modelname on BAGEL~\citep{bagel} and Hunyuan Image 3.0~\citep{hunyuanimage30}, and we refer to these two variants as \modelname-7B and \modelname-80B, respectively.
Both models are trained on a mixed dataset of 5 million samples, combining our curated reasoning-layout-image quadruples, layout-aware editing and understanding datasets, along with additional large-scale text-image pairs to maintain generalization.
For \modelname-7B, we set the batch size of 256 with a learning rate of 2.5e-5. For \modelname-80B, we use a batch size of 1536 and a learning rate of 1e-4.
For the RL-based alignment stage, we use Flow-GRPO with a learning rate of 1e-6, a group size of 16 images per prompt-layout condition, a clipping range of 0.2, and a KL coefficient of 0.01.

Our experiments aim to answer the following questions:
\begin{itemize}[leftmargin=*,itemsep=2pt,topsep=2pt,parsep=2pt]
    \item \textbf{RQ1:} How does \modelname perform on compositional text-to-image generation tasks compared to state-of-the-art baselines?
    \item \textbf{RQ2:} How effective is \modelname at generating layouts that reflect complex spatial instructions?
    \item \textbf{RQ3:} Does \modelname support precise user control through layout editing?
    \item \textbf{RQ4:} Can \modelname effectively handle multimodal understanding and editing tasks?
    \item \textbf{RQ5:} What is the contribution of each key component of \modelname to the overall performance?
\end{itemize}
\label{sec:experiments}

\begin{figure*}
\centering
\includegraphics[width=\textwidth]{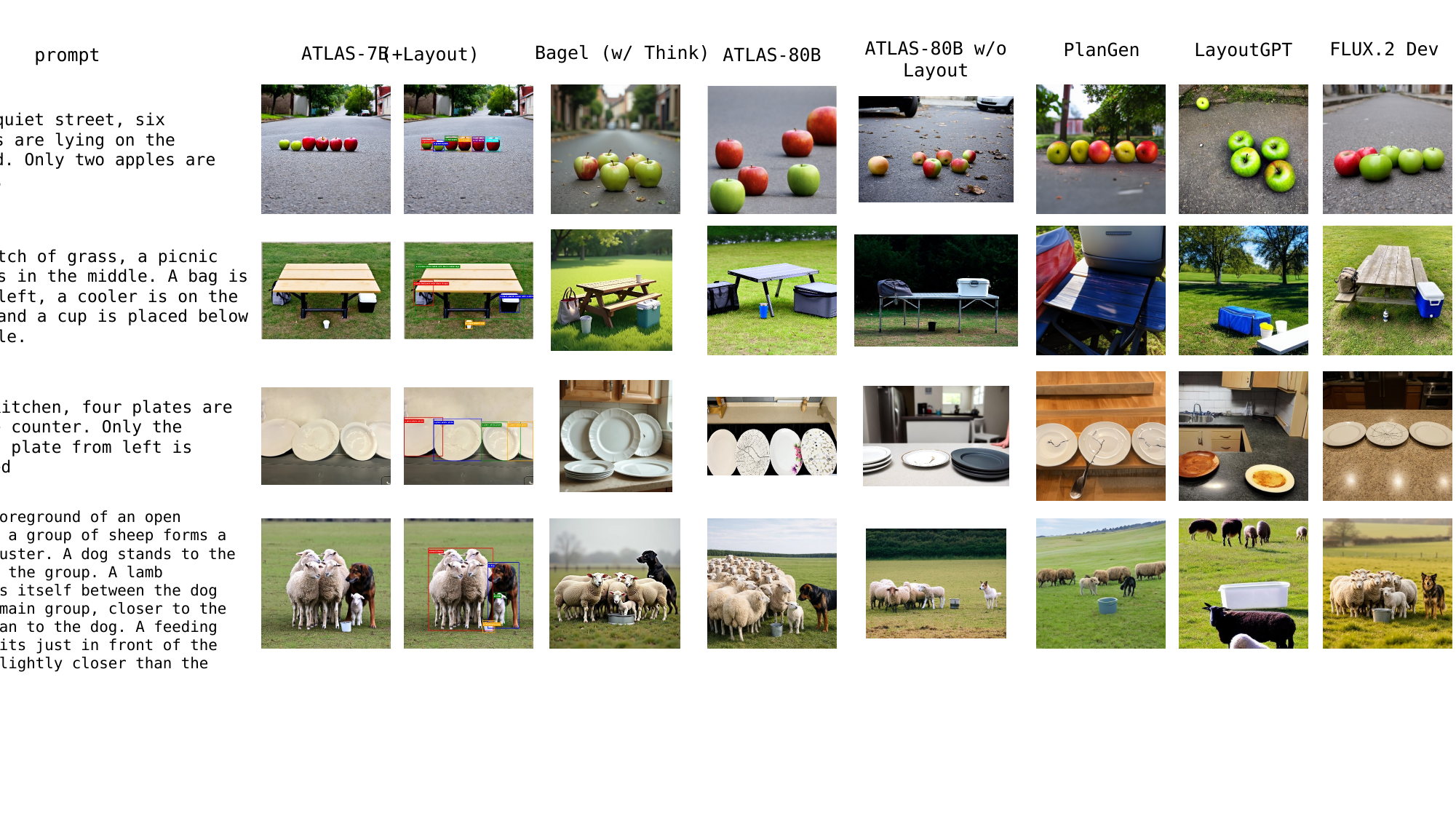}
\caption{Qualitative comparison of text-to-image generation. \modelname better follows complex prompts involving object counts, attributes, and spatial relations.}
\label{fig:t2i_compare}
\end{figure*}
\subsection{RQ1: Text-to-Image Generation Evaluation}

\noindent\textbf{Datasets.} To comprehensively evaluate text-to-image generation capabilities, we utilize two benchmarks:
\begin{itemize}[leftmargin=*,itemsep=0pt,topsep=0pt,parsep=0pt]
    \item \textbf{T2I-CompBench~\citep{t2i-compbench}:} A standardized benchmark for compositional generation, covering dimensions such as color binding, shape, and spatial relationships.
    \item \textbf{\modelname-Reasoning:} To evaluate the reasoning capability of our CoT mechanism, we construct a challenging dataset comprising 500 complex prompts. These prompts require \textbf{multi-step reasoning} to solve, including entangled spatial relations (e.g., ``A is left of B, and B is above C'') or attribute binding (e.g., ``There are four balloons, of which two are red and are not next to each other'').
\end{itemize}
We follow the standard practice in T2I-CompBench to use MLLMs for evaluation. Specifically, we use the prompts provided by T2I-CompBench and employ Qwen3-VL-235B-A22B~\citep{qwen3vl} to assess generation quality. For \modelname-Reasoning, we follow the method for evaluating spatial relations in T2I-CompBench.

\textbf{Results.} 
We compare \modelname against state-of-the-art general text-to-image models, layout-based generation pipelines, and layout-based unified MLLMs. As shown in Table~\ref{tab:main_results}, \modelname-80B achieves results comparable to closed-source models such as Nano Banana Pro on T2I-CompBench, while outperforming layout-based pipelines. On the other hand, ATLAS-7B already surpasses existing layout-based unified MLLMs by 56.59\% on average, demonstrating the effectiveness of our approach across different model scales. Especially on the challenging \modelname-Reasoning benchmark, PlanGen only achieves 25.43, while the two scales of \modelname still maintain strong reasoning capabilities.

Figure~\ref{fig:t2i_compare} presents qualitative comparisons. \modelname outperforms competitors when the prompts involve complex spatial arrangements, or objects have various attributes. For example, when prompted with ``six apples are lying on the ground. Only two apples are green.'' The thinking process of \modelname correctly deduces the number and colors of apples, while the layout concretely specifies their positions and colors. In contrast, other models either miscount the apples or fail to assign the correct colors, leading to inaccurate generations.

\begin{table}
\centering
\caption{\textbf{Layout Planning Evaluation.} Comparison of generated layout quality against baselines. We report both spatial relation scores judged by Qwen3-VL-235B-A22B and rule-based clause checking accuracy on T2I-CompBench (Spatial) and \modelname-Reasoning benchmarks.}
\label{tab:layout_planning}
\resizebox{\linewidth}{!}{%
    \begin{tabular}{l l ccccc}
    \toprule
    \textbf{Eval.} & \textbf{Dataset} & PlanGen~\citep{plangen} & \modelname-7B~w/o Think & \textbf{\modelname-7B~} & \modelname-80B~w/o Think & \textbf{\modelname-80B~} \\ \midrule
    \multirow{2}{*}{Qwen3-VL-235B-A22B} 
    & T2I-CompBench (Spatial) & 37.31 & 67.31 & 82.30 & 83.68 & \textbf{89.45} \\
    & \modelname-Reasoning & 18.41 & 34.56 & 54.72 & 78.81 & \textbf{84.29} \\ \midrule
    \multirow{2}{*}{Rule-based}
    & T2I-CompBench (Spatial) & 29.84 & 54.63 & 72.96 & 73.12 & \textbf{78.35} \\
    & \modelname-Reasoning & 17.92 & 32.41 & 48.67 & 64.38 & \textbf{73.92} \\
    \bottomrule
    \end{tabular}%
}
\end{table}
\begin{figure}
\centering
\includegraphics[width=\linewidth]{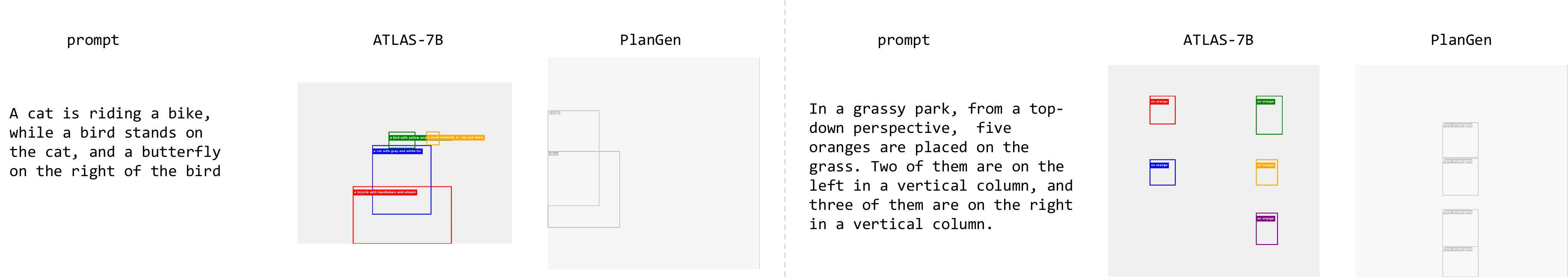}
\caption{Qualitative comparison of layout planning. }
\label{fig:layout_compare}
\end{figure}
\begin{table}
\centering
\caption{\textbf{Layout Shift Consistency on T2I-CompBench.} We report the Hit Rate (HR@0.5) under the challenging Adjacency Shift setting using Qwen3-VL-235B-A22B and GroundingDINO as detectors. A higher score indicates the model can precisely generate objects at new, user-specified locations without positional inertia.}
\label{tab:layout_shift}
\resizebox{0.4\columnwidth}{!}{%
\begin{tabular}{l cc}
\toprule
\textbf{Detector} & PlanGen~\citep{plangen} & \textbf{\modelname-7B} \\ \midrule
Qwen3-VL-235B-A22B & 79.69 & \textbf{90.17}\\
GroundingDINO & 77.34 & \textbf{87.66}\\
\bottomrule
\end{tabular}%
}
\end{table}

\begin{table*}
    \centering
    \caption{Ablation studies on \modelname.}
    
    \label{tab:ablation}
    \resizebox{0.8\linewidth}{!}{%
    \begin{tabular}{l cccccc}
    \toprule
    \multirow{2}{*}{\textbf{Method}} & \multicolumn{6}{c}{\textbf{T2I-CompBench}} \\ 
    \cmidrule(lr){2-7} 
     & Color & Shape & Texture &  Spatial & Non-Spa. & Numeracy  \\ \midrule
    \rowcolor{gray!10} \textbf{Full Model (ATLAS-7B)}  & 97.13 & \textbf{86.17} & \textbf{96.26} & \textbf{86.01} & 93.79 & \textbf{83.41} \\\midrule 
    w/o RL & \textbf{97.41} & 83.31 & 93.35 & 85.83 & 93.33 & 80.83 \\
    w/o Shared Tokens (Plan with text and number) &  95.13	& 79.70 & 94.27 & 81.16 & \textbf{94.81} & 73.02\\
    w/o MMU data & 91.29 & 79.90 & 88.81 &79.83 &91.67 &65.83 \\
    w/o Think & 97.29 & 82.30 & 93.27 & 75.17 & 93.19 & 62.33 \\
    w/o Plan & 92.42 & 80.16 & 90.62 & 80.83 & 91.67 & 61.08 \\
    \bottomrule
    \end{tabular}%
    }
\end{table*}

\begin{table*}[h]
\caption{Evaluation results on Instruction-Guided Editing.}
\label{tab:imgedit}
\label{tab:editing}
    \centering
    \resizebox{0.9\linewidth}{!}{%
        \begin{tabular}{l cccccccccc}
            \toprule
            \multirow{2}{*}{\textbf{Method}} & \multicolumn{10}{c}{\textbf{ImgEdit-Bench}} \\
                & \textbf{Add} & \textbf{Adjust} & \textbf{Extract} & \textbf{Replace} & \textbf{Remove} & \textbf{Background} & \textbf{Style} & \textbf{Hybrid} & \textbf{Action} & \textbf{Overall} \\ \midrule
            BAGEL-7B~\citep{bagel}  & 3.56 & 3.31 & 1.88 & 2.62 & 2.88 & 3.44 & 4.49 & 2.38 & 4.17 & 3.20 \\
            \rowcolor{gray!10} \textbf{\modelname-7B} & \textbf{4.04} & 3.33 & 1.87 & \textbf{3.98} & \textbf{3.21} & \textbf{3.57} & 4.41 & 2.41 & \textbf{4.34} & \textbf{3.46}\\ \bottomrule
        \end{tabular}%
    }
\end{table*}

\begin{table}
    \caption{Evaluation results on Multimodal Understanding. We report RefCOCO, MMMU, and POPE scores.}
    \centering
    \label{tab:mmu}
    \resizebox{0.45\linewidth}{!}{%
        \begin{tabular}{lccc}
            \toprule
            & RefCOCO & MMMU & POPE \\ \midrule
            Qwen2.5-VL-7B~\citep{qwen2vl} & \textbf{91.7} & 53.1 & \textbf{90.2} \\
            \textbf{\modelname-7B} & 90.5 & \textbf{54.7} & 88.1 \\ \bottomrule
        \end{tabular}%
    }
\end{table}

\subsection{RQ2: Layout Planning Evaluation}

\label{sec:rq2_control}
To assess layout planning capabilities of \modelname, we evaluate the quality of generated bounding boxes given text prompts. We compare \modelname against PlanGen and \modelname without the \textbf{``Think''} phase (i.e., directly generating layouts from text). We use the \modelname-Reasoning dataset and the spatial subset of T2I-CompBench for evaluation.
We evaluate layouts with Qwen3-VL-235B-A22B~\citep{qwen3vl} and a rule-based clause checker, measuring whether each object is present and each spatial relation is satisfied.

\textbf{Results.}
Table \ref{tab:layout_planning} summarizes the results under both evaluators. \modelname outperforms PlanGen by a large margin, and the inclusion of the \textbf{``Think''} phase significantly boosts layout planning accuracy. This confirms that the ``Think'' phase effectively enhances the model's reasoning about spatial relationships before layout generation.
Under Qwen3-VL-235B-A22B evaluation, the ``Think'' phase brings larger improvements on the smaller model (7B), improving the score by an average of 17.6, compared with 5.62 on the 80B model. Rule-based checking shows the same trend. We attribute this to the fact that the larger model's ability is already strong enough to handle most spatial arrangements without explicit CoT.

Figure~\ref{fig:layout_compare} shows qualitative examples. The ``Think'' phase enables \modelname to generate layouts that accurately reflect complex spatial instructions, while PlanGen often produces layouts with missing objects, incorrect quantifiers, and spatial relations.

\subsection{RQ3: Precision in Controllable Generation}
\label{sec:rq3_control}

To address RQ3, we evaluate the fidelity of the ``Plan $\to$ Paint'' phase. In real-world interactive applications, users typically utilize bounding boxes as precise control handles to refine generated results. Therefore, a truly controllable model must be responsive to these manual adjustments, strictly adhering to the new spatial constraints rather than relying on internal priors or ``positional inertia'' from the initial generation.

We simulate this interactive refinement scenario using the T2I-CompBench dataset. Specifically, we conduct a controlled Layout Shift Test on valid layouts where the target object is correctly generated and matches the original box. For this target object, we apply an Adjacency Shift perturbation: we spatially translate the bounding box until it is disjoint from but adjacent to its original position (sharing only a border, resulting in an $\text{IoU} \approx 0$). This setting represents a challenging limit case, as it forces the model to synthesize the object in a completely fresh region while effectively ``erasing'' it from the original coordinates.

We regenerate the image conditioned on the \textit{shifted} layout and detect the target object with Qwen3-VL-235B-A22B and GroundingDINO~\citep{groundingdino}. We report the Hit Rate (HR@0.5), which counts a generation as successful if the detected object overlaps with the \textit{new} target box with an IoU $> 0.5$. This metric strictly measures the execution precision of the spatial adjustment.

\textbf{Results.} As shown in Table~\ref{tab:layout_shift}, \modelname~achieves the highest success rate under both detectors. While the baseline method often struggles to relocate objects away from their initial or canonical positions (positional inertia), our method effectively adheres to the updated layout constraints, validating its superior controllability for interactive fine-grained layout adjustments.

\subsection{RQ 4: Versatility of the Unified Framework}

To address \textbf{RQ4}, we demonstrate that \modelname~is not merely a generator, but a unified system capable of precise editing and multimodal understanding.

\noindent\textbf{Instruction-Guided Editing.}
We evaluate the editing capabilities on the ImgEdit-Bench~\citep{imgedit} dataset.
We compare \modelname-7B against its base model, BAGEL-7B, to verify the contribution of our unified planning mechanism.
As shown in Table~\ref{tab:editing}, \modelname~achieves consistent improvements over BAGEL on average.
Especially on Add, Replace, and Remove tasks, \modelname achieves significant improvements.
This indicates that explicitly modeling the "Plan" phase allows for more precise structural manipulation.

\noindent\textbf{Multimodal Understanding.}
We further evaluate multimodal understanding on RefCOCO grounding, MMMU~\citep{mmmu}, and POPE~\citep{pope}, comparing \modelname-7B with Qwen2.5-VL-7B~\citep{qwen2vl}.
As shown in Table~\ref{tab:mmu}, \modelname-7B remains competitive on RefCOCO and POPE, while slightly outperforming Qwen2.5-VL-7B on MMMU.

\subsection{RQ 5: Ablation Studies}
\label{sec:ablation}

To answer \textbf{RQ5} and quantify the contribution of each core component, we conduct ablation studies on the T2I-Compbench benchmark. We train several variants of \modelname-7B by selectively removing key modules or training strategies: RL-based layout alignment, shared positional tokens, multimodal understanding data, the \textbf{``Think''} phase, and the \textbf{``Plan''} phase. As shown in Table~\ref{tab:ablation}, removing these components leads to performance drops, especially on spatial and numeracy dimensions, which highlights the importance of these designs in aligning text and vision modalities for precise spatial reasoning and layout planning.
Additionally, the results discussed in Section~\ref{sec:rq2_control} can further validate the effectiveness of the ``Think'' phase.

\section{Conclusion}
In this paper, we presented ATLAS, a unified native MLLM framework that redefines controllable image synthesis through a human-like ``Think, Plan, Paint'' paradigm. By introducing a Layout-Aware Chain-of-Thought mechanism and Shared Positional Tokens, ATLAS effectively bridges the semantic gap between abstract textual reasoning and concrete visual planning. Extensive evaluations across 7B and 80B scales demonstrate that our approach not only establishes new state-of-the-art results on complex compositional generation benchmarks but also achieves versatile performance in instruction-guided editing and visual grounding. Ultimately, ATLAS offers a scalable, transparent, and robust path toward fine-grained controllable image generation in unified multimodal models.

\nocite{*}

\bibliographystyle{plainnat}
\bibliography{atlas.bib}

\newpage
\appendix
\section{LLM Usage Statements}
We used LLMs to polish our writing during paper preparation. 

\section{Limitations}
\label{appendix:limitations}
This work uses bounding boxes as the layout interface. This representation is simple, interpretable, and well aligned with existing grounding and layout-aware generation benchmarks, which makes it suitable for studying the alignment between textual reasoning, layout planning, and visual generation. 

At the same time, bounding boxes are only one possible way to represent layouts. Other interfaces may better capture certain types of visual structure or user intent, especially in scenarios that require more detailed spatial control. Exploring how different layout interfaces affect reasoning, planning, and generation quality is an important direction for future work.

\section{Broader Impacts}
\label{appendix:broader_impacts}

This work aims to improve controllable image generation by enabling models to better follow explicit layout instructions. Such capability can benefit creative design, visual communication, education, and accessibility-oriented content creation, where users may need more precise control over object placement and spatial relationships.

At the same time, more controllable image generation can also lower the barrier to producing misleading or harmful visual content. Potential misuse includes generating deceptive media, manipulating visual narratives, or creating biased and inappropriate images. We encourage responsible deployment with content provenance, safety filtering, and human oversight, especially in applications that may affect public information, identity, or vulnerable groups.

\section{Experimental Details}
\label{appendix:exp_details}

\subsection{Compute Resources}
\label{appendix:compute_resources}

Our experiments were conducted on a GPU cluster with 256 GPUs in total, where each GPU is equipped with 96GB of memory.

\subsection{Statistical Significance Tests}
\label{appendix:significance_tests}

For the main text-to-image generation results, we conduct paired significance tests over prompt-level evaluation scores.
We compare ATLAS with PlanGen and with the corresponding base model, i.e., BAGEL for ATLAS-7B and Hunyuan Image 3.0 for ATLAS-80B.
Using paired bootstrap resampling, we test whether the mean prompt-level score difference is greater than zero.
At the significance level $\alpha=0.01$, the improvements of ATLAS remain statistically significant in both comparison groups.

\subsection{Reward Details for RL-based Layout Alignment}
\label{appendix:rl_rewards}

We use five rewards in the RL-based layout alignment stage and combine them with a weighted sum:
\begin{equation}
    r(I,T,L) =
    w_{\mathrm{pos}} r_{\mathrm{pos}}
    + w_{\mathrm{sem}} r_{\mathrm{sem}}
    + w_{\mathrm{cnt}} r_{\mathrm{cnt}}
    + w_{\mathrm{neg}} r_{\mathrm{neg}}
    + w_{\mathrm{qua}} r_{\mathrm{qua}} .
\end{equation}

\textbf{Layout-position reward.}
We estimate object boxes from the generated image and match them with the input layout boxes using semantic labels and spatial overlap.
The reward is computed as an IoU-based F1 score:
\begin{equation}
    r_{\mathrm{pos}} =
    \frac{2 \cdot \mathrm{Precision} \cdot \mathrm{Recall}}
    {\mathrm{Precision} + \mathrm{Recall} + \epsilon}.
\end{equation}

\textbf{Region-counting reward.}
For a layout with $N$ regions, let $c_j$ be the expected number of target objects in the $j$-th region, and let $\hat{c}_j$ be the estimated count from the generated image.
We define:
\begin{equation}
    r_{\mathrm{cnt}} =
    \frac{1}{N}\sum_{j=1}^{N}
    \max\left(0, 1 - \frac{|\hat{c}_j-c_j|}{\max(c_j,1)}\right).
\end{equation}

\textbf{Semantic-alignment reward.}
To preserve global prompt fidelity, we compute the CLIP similarity between the text prompt and the generated image and normalize it into a bounded reward:
\begin{equation}
    r_{\mathrm{sem}} =
    \sigma\left(\tau \cdot
    \mathrm{cos}\left(E_{\mathrm{text}}(T), E_{\mathrm{image}}(I)\right)\right),
\end{equation}
where $E_{\mathrm{text}}$ and $E_{\mathrm{image}}$ are the text and image encoders, and $\tau$ is a temperature.

\textbf{Negative-region reward.}
This reward penalizes target objects that appear outside the specified layout regions.
Given the number of unmatched detected target objects $N_{\mathrm{unmatched}}$ and all detected target objects $N_{\mathrm{pred}}$, we define:
\begin{equation}
    r_{\mathrm{neg}} =
    1 - \frac{N_{\mathrm{unmatched}}}{\max(N_{\mathrm{pred}},1)}.
\end{equation}

\textbf{Image-quality reward.}
Finally, we use an image preference model to encourage visual realism and aesthetics:
\begin{equation}
    r_{\mathrm{qua}} =
    \sigma\left(s_{\mathrm{quality}}(I)\right).
\end{equation}
All reward terms are normalized to comparable ranges before weighted aggregation.

\section{T2I CompBench Evaluation Details}
\label{sec:eval_prompts}

In our evaluation pipeline, we utilize MLLMs to assess the alignment between generated images and text prompts. Below, we detail the specific instruction templates used for each category in T2I-CompBench. In these templates, placeholders such as \texttt{[Category]}, \texttt{[Object]}, and \texttt{[Prompt]} are dynamically replaced with the specific attributes, objects, or captions from the dataset.

\subsection{Attribute Binding (Color, Shape, Texture)}
For attribute binding tasks, we evaluate whether a specific object possesses the correct attribute. The scoring range is 1-4.

\vspace{0.2cm}
\noindent\fbox{%
    \begin{minipage}{0.95\linewidth}
    \textbf{System Instruction:} You are my assistant to identify any objects and their \texttt{[Category]} in the image. According to the image, evaluate if there is a \texttt{[Object]} in the image.
    
    \textbf{Criteria:}
    \begin{itemize}[leftmargin=1.5em, topsep=0pt, itemsep=0pt]
        \item \textbf{4}: There is \texttt{[Object]}, and \texttt{[Category]} is \texttt{[Attribute]}.
        \item \textbf{3}: There is \texttt{[Object]}, \texttt{[Category]} is mostly \texttt{[Attribute]}.
        \item \textbf{2}: There is \texttt{[Object]}, but it is not \texttt{[Attribute]}.
        \item \textbf{1}: No \texttt{[Object]} in the image.
    \end{itemize}
    
    \textbf{Output Format:} Provide your analysis and explanation in JSON format with the following keys: score (range 1-4, e.g., 1), explanation (within 20 words).
    \end{minipage}%
}

\subsection{Spatial Relationships (Spatial, 3D, Shape-Multistep)}
For spatial categories, we assess the layout accuracy. The scoring range is 1-5.

\vspace{0.2cm}
\noindent\fbox{%
    \begin{minipage}{0.95\linewidth}
    \textbf{System Instruction:} You are my assistant to identify objects and their spatial layout in the image. According to the image, evaluate if the text ``\texttt{[Prompt Text]}'' is correctly portrayed in the image.
    
    \textbf{Criteria:}
    \begin{itemize}[leftmargin=1.5em, topsep=0pt, itemsep=0pt]
        \item \textbf{5}: Correct spatial layout in the image for all objects mentioned in the text.
        \item \textbf{4}: Basically, spatial layout of objects matches the text.
        \item \textbf{3}: Spatial layout not aligned properly with the text.
        \item \textbf{2}: Image not aligned properly with the text.
        \item \textbf{1}: Image almost irrelevant to the text.
    \end{itemize}
    
    \textbf{Output Format:} Provide your analysis and explanation in JSON format with the following keys: score (range 1-5, e.g., 2), explanation (within 50 words).
    \end{minipage}%
}

\subsection{Action and Non-Spatial Relationships}
This category evaluates actions, events, and inter-object relationships. The scoring range is 1-5.

\vspace{0.2cm}
\noindent\fbox{%
    \begin{minipage}{0.95\linewidth}
    \textbf{System Instruction:} You are my assistant to identify the actions, events, objects and their relationships in the image. According to the image, evaluate if the text ``\texttt{[Prompt Text]}'' is correctly portrayed in the image.
    
    \textbf{Criteria:}
    \begin{itemize}[leftmargin=1.5em, topsep=0pt, itemsep=0pt]
        \item \textbf{5}: The image accurately portrayed the actions, events and relationships between objects described in the text.
        \item \textbf{4}: The image portrayed most of the actions, events and relationships but with minor discrepancies.
        \item \textbf{3}: The image depicted some elements, but action relationships between objects are not correct.
        \item \textbf{2}: The image failed to convey the full scope of the text.
        \item \textbf{1}: The image did not depict any actions or events that match the text.
    \end{itemize}
    
    \textbf{Output Format:} Provide your analysis and explanation in JSON format with the following keys: score (range 1-5, e.g., 2), explanation (within 20 words).
    \end{minipage}%
}

\subsection{Numeracy}
This category evaluates the correct quantity of objects. The scoring range is 1-5.

\vspace{0.2cm}
\noindent\fbox{%
    \begin{minipage}{0.95\linewidth}
    \textbf{System Instruction:} You are my assistant to identify objects and their quantities in the image. According to the image and your previous answer, evaluate how well the image aligns with the text prompt: ``\texttt{[Prompt Text]}''.
    
    \textbf{Criteria:}
    \begin{itemize}[leftmargin=1.5em, topsep=0pt, itemsep=0pt]
        \item \textbf{5}: Correct numerical content in the image for all objects mentioned in the text.
        \item \textbf{4}: Basically, numerical content of objects matches the text.
        \item \textbf{3}: Numerical content not aligned properly with the text.
        \item \textbf{2}: Image not aligned properly with the text.
        \item \textbf{1}: Image almost irrelevant to the text.
    \end{itemize}
    
    \textbf{Output Format:} Provide your analysis and explanation in JSON format with the following keys: score (range 1-5, e.g., 2), explanation (within 20 words).
    \end{minipage}%
}

\subsection{Additional Results}

\subsubsection{Editing Examples}

\begin{figure}
\centering
\includegraphics[width=0.9\textwidth]{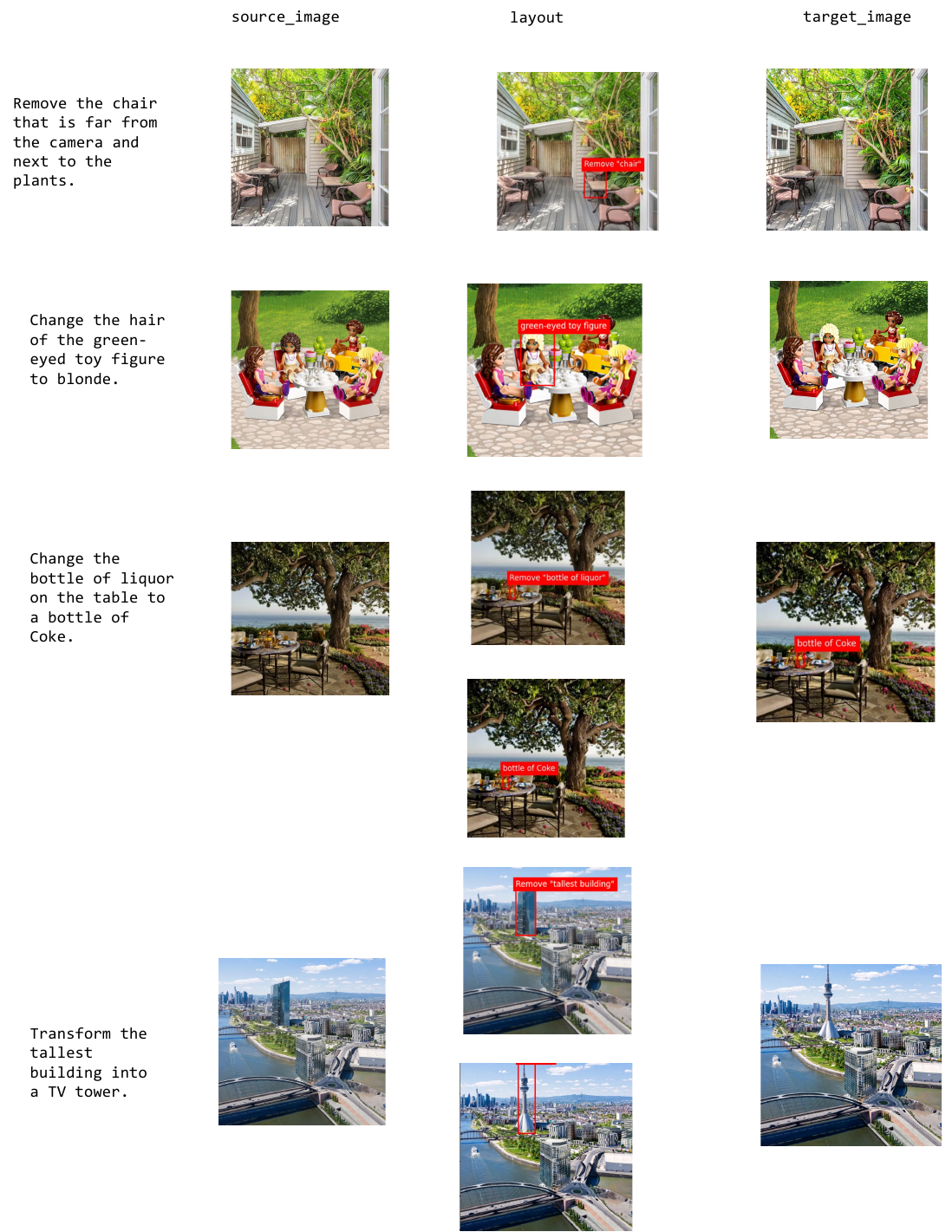}

\caption{Additional editing examples using \modelname.}
\label{fig:editing_examples}
\end{figure}

We further present editing examples in Figure~\ref{fig:editing_examples}, showcasing \modelname's ability to adapt to user-modified layouts and generate images that accurately reflect these changes.
Notably, we do not need to manually use negative prompts to prevent the presence of removed objects, as \modelname inherently understands and adheres to the updated layout instructions.

\subsubsection{Layout Diversity}

\begin{figure}
\centering
\includegraphics[width=\textwidth]{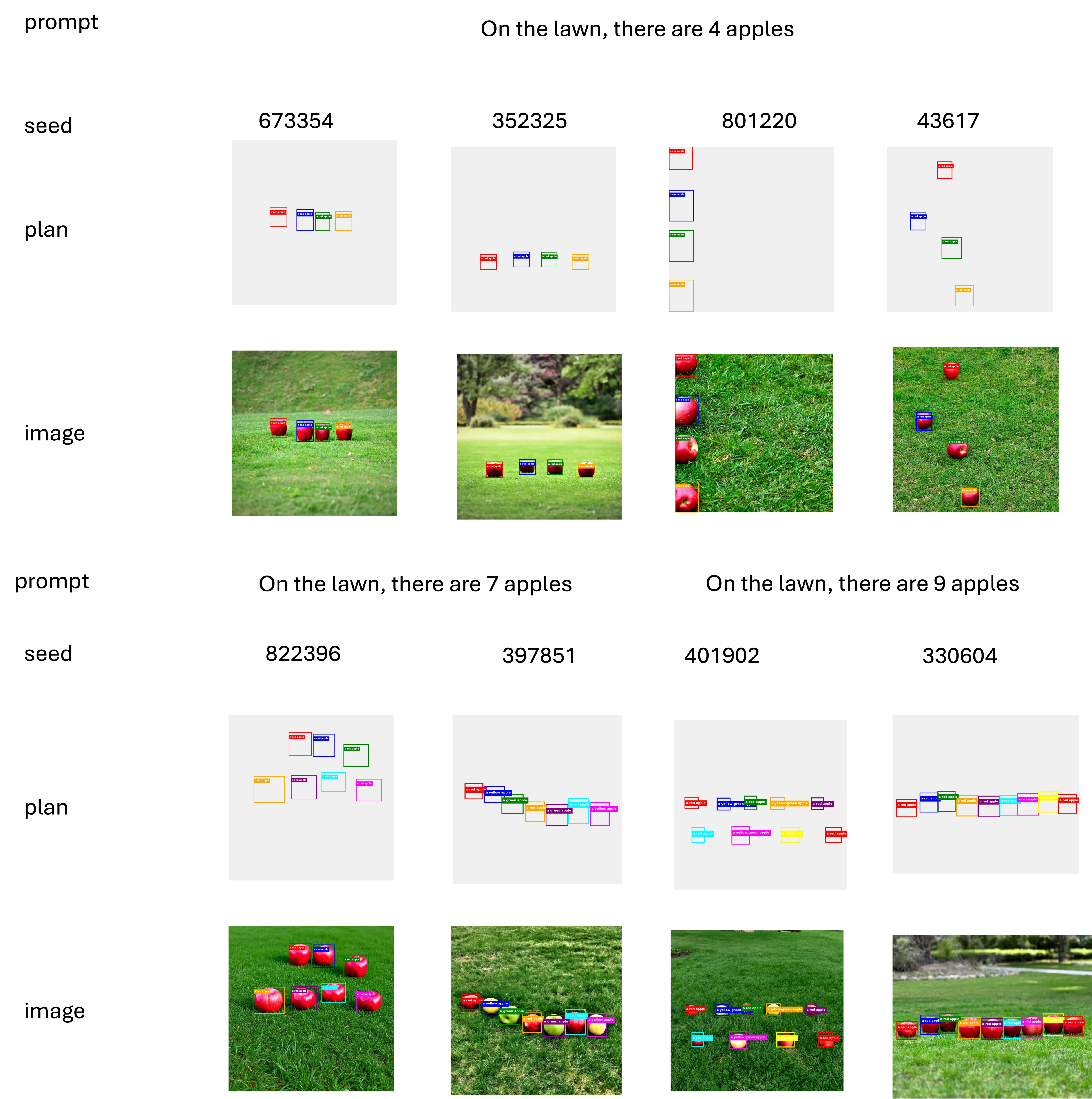}
\caption{Layout diversity under different random seeds. For the same counting prompts, \modelname samples multiple valid layouts with distinct object positions and optional attributes, while preserving the target object counts in the generated images.}
\label{fig:layout_diversity}
\end{figure}

A potential concern for layout-based generation is that the intermediate layout representation may collapse to a small set of fixed spatial templates.
\modelname avoids this issue because both the Think and Plan stages are generated autoregressively by the underlying LLM rather than retrieved from deterministic templates.
As a result, standard stochastic decoding naturally induces a distribution over plausible layouts conditioned on the same text prompt.
As shown in Figure~\ref{fig:layout_diversity}, we qualitatively examine this property by sampling multiple layouts for counting-specific prompts with different random seeds, including prompts that require 4, 7, and 9 target objects.
Across these samples, \modelname preserves the requested object counts while producing diverse spatial arrangements and corresponding generated images.
When the prompt leaves certain attributes unspecified, the sampled layouts can also vary optional object attributes such as color.
This suggests that explicit layout planning provides controllability without sacrificing diversity, allowing the model to represent multiple valid visual realizations for the same instruction.

\subsubsection{Ambiguous Spatial Language}

\begin{figure}
\centering
\includegraphics[width=0.75\textwidth]{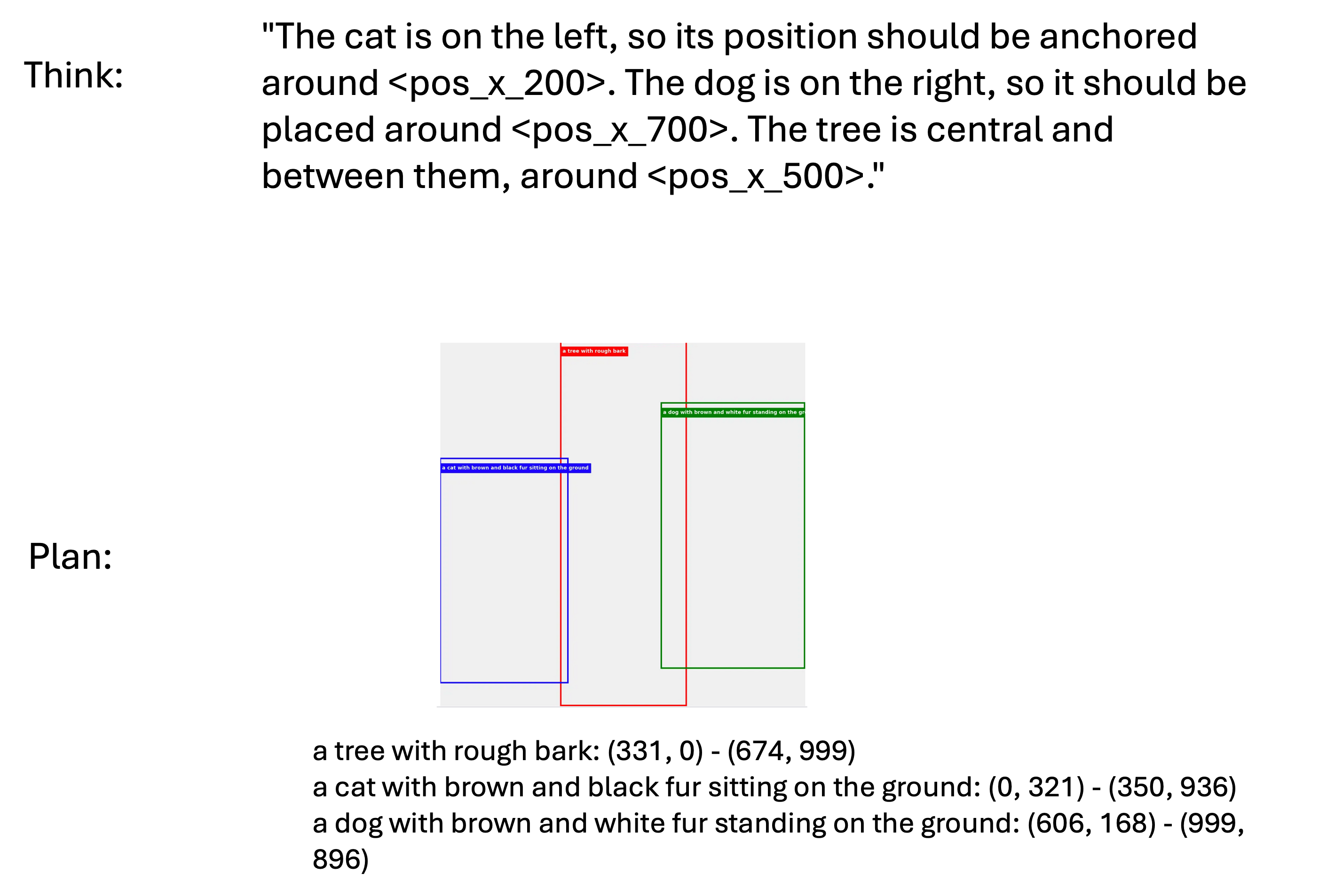}
\caption{Handling ambiguous spatial language. For prompts such as ``a cat and a dog around a tree,'' \modelname first produces a reasonable default layout that satisfies the underspecified relation.}
\label{fig:ambiguous_near_around}
\end{figure}

\begin{figure}
\centering
\includegraphics[width=0.9\textwidth]{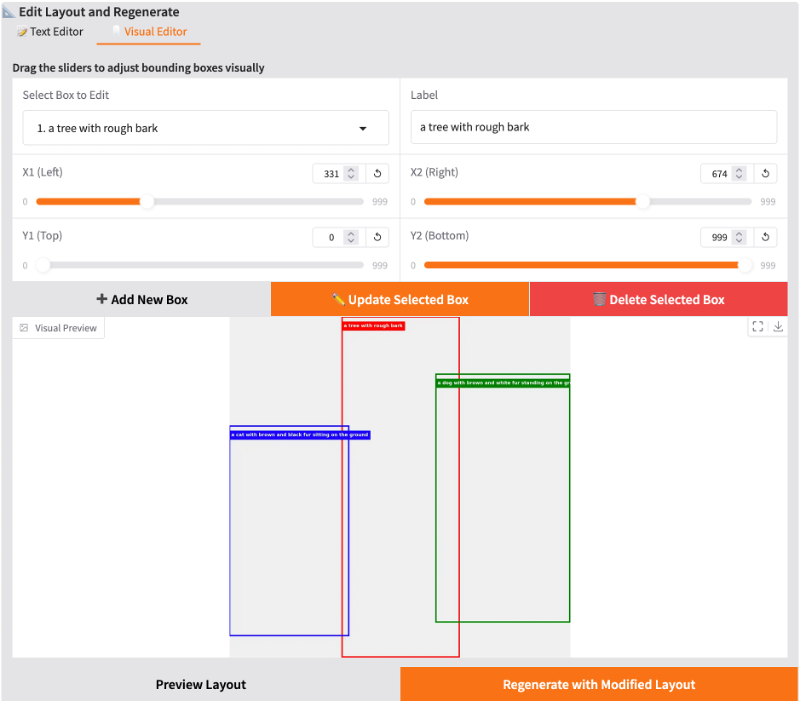}
\caption{Interactive plan editing interface. Users can adjust, add, or delete intermediate bounding boxes before regeneration, which allows ambiguous spatial instructions to be refined into a user-specified arrangement.}
\label{fig:plan_edit_interface}
\end{figure}

Natural language spatial terms such as ``near'' and ``around'' are inherently ambiguous, since they often correspond to multiple valid object arrangements rather than a single canonical layout.
\modelname handles such prompts by combining natural-language reasoning with an editable intermediate plan.
As shown in Figure~\ref{fig:ambiguous_near_around}, for the prompt ``a cat and a dog around a tree,'' the Think stage explicitly reasons about the spatial relation and the Plan stage instantiates a physically and compositionally reasonable default layout.
This default should be viewed as one plausible interpretation of the underspecified prompt, not as the only valid arrangement.
When users have a more specific configuration in mind, they can directly modify the intermediate coordinates before generation, as illustrated in Figure~\ref{fig:plan_edit_interface}.
This turns ambiguity into an interactive control opportunity: \modelname provides a reasonable initial plan, while the user can refine the layout before the final Paint stage.

\subsubsection{Reasoning--Generation Interaction}

\begin{figure}
\centering
\includegraphics[width=0.9\textwidth]{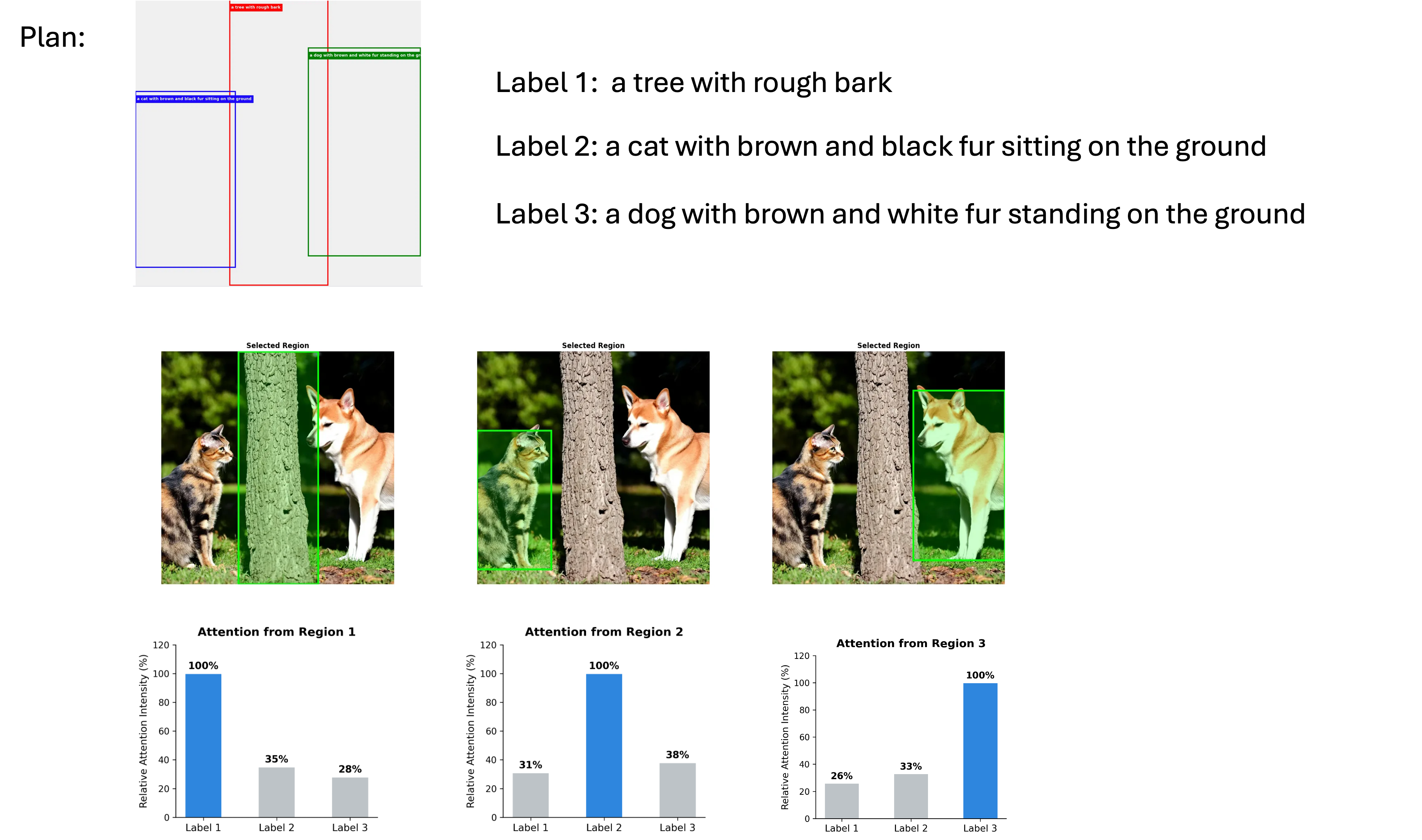}
\caption{Attention analysis between layout planning and image generation. For each selected generated region, we visualize the attention from image tokens back to object labels in the layout plan. The generated regions attend most strongly to their corresponding layout labels, indicating that the Paint stage follows the object-level plan produced by the reasoning stage.}
\label{fig:reasoning_generation_attention}
\end{figure}

The Paint stage is strictly conditioned on the output sequence produced by the preceding Think and Plan stages.
To illustrate how the layout plan guides image synthesis, we analyze attention weights from generated image regions back to the object labels in the layout plan.
As shown in Figure~\ref{fig:reasoning_generation_attention}, each selected region in the generated image attends most strongly to its corresponding object label in the plan.
For example, image tokens in the cat, tree, and dog regions assign the highest relative attention to the matching layout labels rather than unrelated objects.
This indicates that the generation process does not merely use the layout as a loose global hint; instead, it maintains object-level correspondence between planned regions and generated visual content, helping the model focus on the areas specified by the intermediate plan.

\subsubsection{Beyond Spatial Reasoning}

\begin{figure}
\centering
\includegraphics[width=0.9\textwidth]{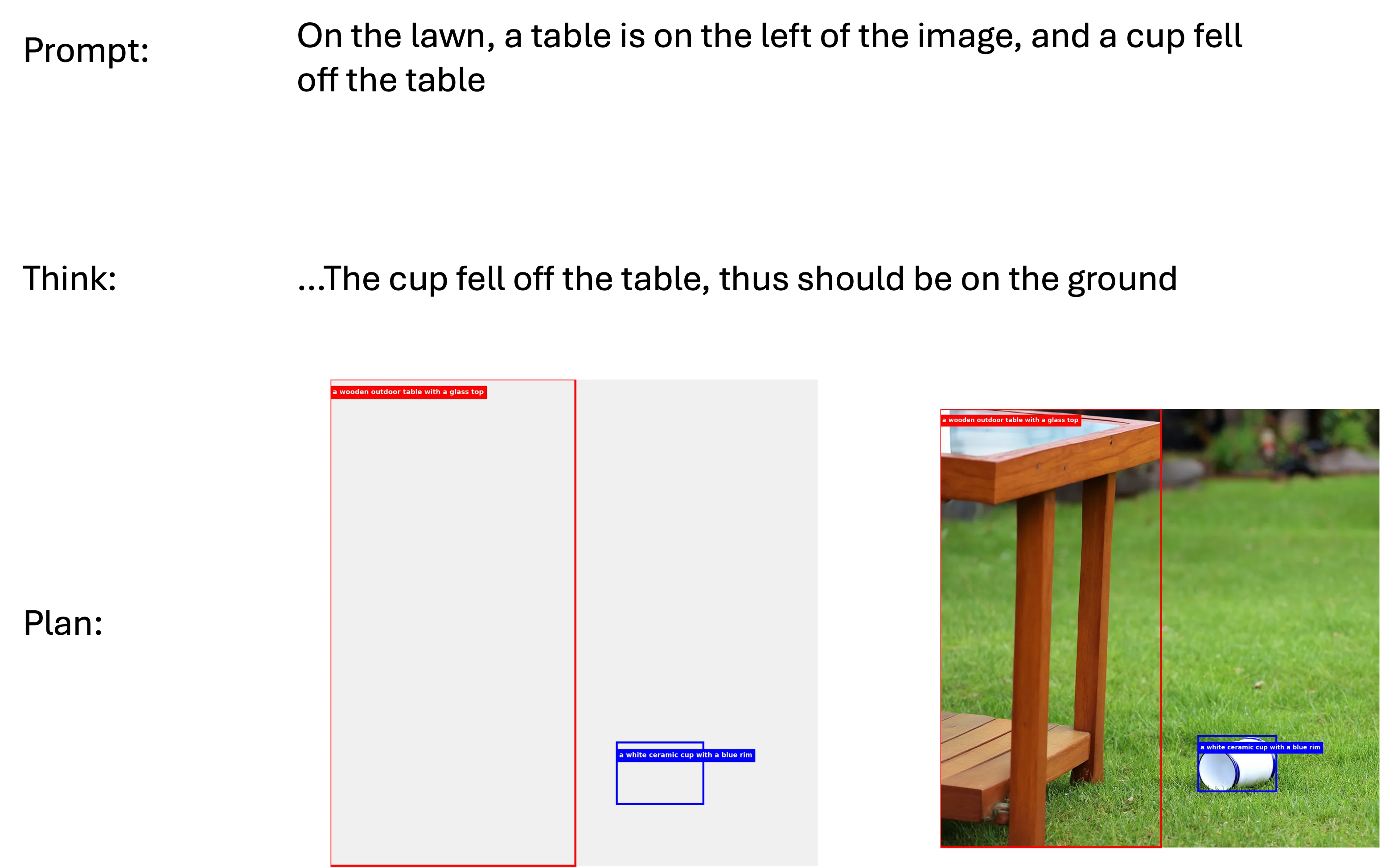}
\caption{Example of physical reasoning in layout planning. Given a prompt implying that a cup has fallen off a table, \modelname infers in the Think stage that the cup should be placed on the ground rather than on the tabletop, and the generated image follows this physically plausible layout.}
\label{fig:physical_causal_reasoning}
\end{figure}

Although \modelname uses layouts as the intermediate representation, the reasoning performed before layout generation is not limited to explicit spatial relations.
Because the Think stage is generated in the text modality, it can incorporate broader commonsense knowledge, including physical and causal implications that are only implicit in the prompt.
Figure~\ref{fig:physical_causal_reasoning} shows an example with the prompt ``a cup fell off the table.''
Rather than placing the cup on the tabletop, \modelname infers that the cup should appear on the ground after falling, and the subsequent Plan and Paint stages realize this physically plausible scene.
This example illustrates that layout planning can serve as an interface for grounding not only spatial instructions, but also higher-level reasoning about events and their visual consequences.

\subsubsection{Text-to-Image Generation Benchmarks}

\begin{table}
    \centering
    \caption{DPG-Bench evaluation results.}
    \label{tab:dpg_bench}
    \begin{tabular}{lcccccc}
      \toprule
      Model & Global & Entity & Attribute & Relation & Other & Overall$\uparrow$ \\
      \midrule
      SDXL & 83.27 & 82.43 & 80.91 & 86.76 & 80.41 & 74.65 \\
      BAGEL & 88.94 & 90.37 & 91.29 & 90.82 & 88.67 & 85.07 \\
      BAGEL (Think) & 87.29 & 88.67 & 91.30 & 91.67 & 87.11 & 85.92 \\
      ATLAS-80B w/o Layout & 90.17 & 89.36 & 90.53 &88.42 & 89.21 & 86.02 \\
      ATLAS-80B w/o Layout (Think) & 89.62 & 90.01 & 90.28 & 91.96 & 89.72 & 86.78 \\
      PlanGen & 87.58 & 88.63 & 88.17 & 91.30 & 88.30 & 85.63 \\
      \rowcolor{gray!10} \textbf{ATLAS-7B} & \textbf{90.41} & 90.13 & \textbf{92.51} & 92.75 & 89.54 & 87.29 \\
      \rowcolor{gray!10} ATLAS-80B & 90.13 & \textbf{91.36} & 92.43 & \textbf{94.42} & \textbf{90.32} & \textbf{88.20} \\
      \bottomrule
    \end{tabular}
\end{table}

\paragraph{DPG-Bench.}
Table~\ref{tab:dpg_bench} reports the detailed DPG-Bench results, which evaluate prompt following from multiple perspectives including global consistency, entity coverage, attribute binding, and relations.
\modelname achieves the best overall performance among all compared methods, with \modelname-80B reaching an overall score of 88.20 and \modelname-7B reaching 87.29.
Compared with the corresponding base models, the two \modelname variants consistently improve overall prompt-following accuracy, demonstrating that explicit layout-aware reasoning benefits not only spatial control but also general compositional alignment.
The gains are especially clear on relation-centric and attribute-centric dimensions: \modelname-80B obtains the best relation score of 94.42, while \modelname-7B achieves the best attribute score of 92.51.
These results suggest that the Think--Plan--Paint formulation helps the model bind objects, attributes, and relations into a coherent intermediate layout before image synthesis.

\begin{table}
    \centering
    \caption{GenEval evaluation results.}
    \label{tab:geneval}
    \resizebox{\linewidth}{!}{%
    \begin{tabular}{lccccccc}
      \toprule
      Model & Single Obj. & Two Obj. & Counting & Colors & Position & Color Attri. & Overall$\uparrow$ \\
      \midrule
      SDXL & 0.98 & 0.74 & 0.39 & 0.85 & 0.15 & 0.23 & 0.55 \\
      BAGEL & 0.98 & 0.95 & 0.84 & \textbf{0.95} & 0.78 & \textbf{0.77} & 0.88 \\
      Hunyuan Image 3.0 & \textbf{1.00} & 0.93 & 0.85 & 0.91 & 0.77 & 0.68 & 0.86 \\
      PlanGen &0.98 & 0.81 & 0.51 & 0.89 & 0.66 & 0.59 & 0.72 \\
      \rowcolor{gray!10} ATLAS-7B & 0.98 & \textbf{0.96} & 0.88 & \textbf{0.95} & 0.85 & \textbf{0.77} & 0.90 \\
      \rowcolor{gray!10} ATLAS-80B & 0.99 & \textbf{0.96} & \textbf{0.92} & \textbf{0.95} & \textbf{0.91} & 0.76 & \textbf{0.92} \\
      \bottomrule
    \end{tabular}
    }
\end{table}

\paragraph{GenEval.}
Table~\ref{tab:geneval} further evaluates object-level compositional generation on GenEval.
\modelname-80B achieves the highest overall score of 0.92, and \modelname-7B also reaches a strong overall score of 0.90, outperforming both their base models and the layout-planning baseline.
The improvement is most pronounced on categories that require precise structural grounding, such as counting and position.
For example, \modelname-80B improves the position score from 0.77 to 0.91 over Hunyuan Image 3.0, while \modelname-7B improves the position score from 0.78 to 0.85 over BAGEL.
The consistent gains on counting and spatial placement indicate that explicit layout planning provides a reliable bridge between textual compositional constraints and visual realization.

\subsubsection{Efficiency and Evaluation Details}

\begin{table}
    \centering
    \caption{Runtime comparison of different models. Runtime is measured in seconds per image.}
    \label{tab:runtime}
    \begin{tabular}{lc}
      \toprule
      Model & Runtime (s) \\
      \midrule
      BAGEL & 87.3 \\
      BAGEL (Think) & 98.7 \\
      ATLAS-7B & 99.4 \\
      ATLAS-80B w/o Layout & 143.9 \\
      ATLAS-80B w/o Layout (Think) & 155.2 \\
      ATLAS-80B & 156.5 \\
      \bottomrule
    \end{tabular}
\end{table}

\paragraph{Runtime.}
Table~\ref{tab:runtime} reports the inference latency of \modelname and the corresponding base models, measured in seconds per image under the same evaluation setting.
\modelname introduces only a modest computational overhead over the base generators.
For the 7B model, \modelname-7B takes 99.4 seconds per image, which is close to BAGEL with the Think stage enabled (98.7 seconds) and about 13.9\% slower than the original BAGEL model without thinking.
For the 80B model, \modelname-80B takes 156.5 seconds per image, similarly matching Hunyuan Image 3.0 with thinking (155.2 seconds) and adding about 8.8\% latency over the original Hunyuan Image 3.0.
These results indicate that the explicit Think--Plan--Paint process substantially improves controllability and compositional alignment while keeping the inference cost comparable to base models with reasoning enabled.

\begin{figure}
\centering
\includegraphics[width=\textwidth]{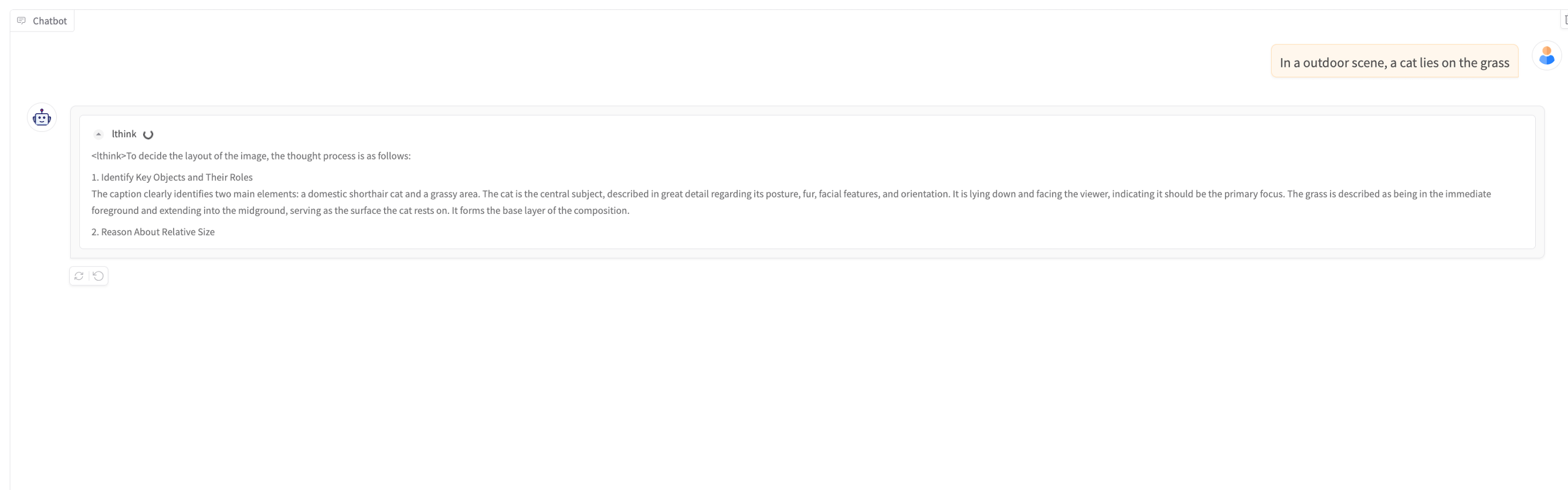}
\caption{Interactive interface for \modelname. The interface streams the model's thinking process to the user and supports interruption during reasoning, enabling more transparent and controllable interaction.}
\label{fig:interactive_usability}
\end{figure}

In interactive scenarios, the thinking stage does not need to be treated as an opaque blocking step.
As shown in Figure~\ref{fig:interactive_usability}, our interface streams the model's thinking process while it reasons about the prompt and prepares the layout plan.
Users can inspect this partial reasoning in real time and interrupt the process at any moment, for example when the streamed reasoning already reveals an undesired interpretation or when they want to revise the prompt before image synthesis.
This design improves practical responsiveness: although the full Think--Plan--Paint pipeline has a modest overhead in end-to-end latency, users are not forced to wait for the entire thinking process when early feedback is sufficient.

\end{document}